\documentclass[conference]{IEEEtran}
\IEEEoverridecommandlockouts
\usepackage{hyperref}
\usepackage{amsmath,amssymb,amsfonts}
\usepackage{algorithmic}
\usepackage{array}
\usepackage{graphicx}
\usepackage{textcomp}
\usepackage[table]{xcolor}
\def\BibTeX{{\rm B\kern-.05em{\sc i\kern-.025em b}\kern-.08em
    T\kern-.1667em\lower.7ex\hbox{E}\kern-.125emX}}
    
\usepackage{amssymb}%
\usepackage{pifont}%
\newcommand{\cmark}{\ding{51}}%
\newcommand{\xmark}{\ding{55}}%

\usepackage{textcomp}

\usepackage{svg}
    
\usepackage{multirow} 
\usepackage{booktabs} %
\definecolor{lightblue}{rgb}{0.93,0.95,1.0}
\usepackage{siunitx}
\usepackage{etoolbox}           %
\renewcommand{\bfseries}{\fontseries{b}\selectfont} %
\robustify\bfseries             %
\newrobustcmd{\B}{\bfseries}    %

\usepackage{afterpage}
\usepackage{xspace}

\usepackage{etoolbox}
\newbool{umap_to_appendix}
\booltrue{umap_to_appendix}
\boolfalse{umap_to_appendix}

\newcommand{\x}{\mathbf{x}}

\newcommand{\y}{\mathbf{y}}
\newcommand{\ye}{\mathbf{\tilde{y}}}
\renewcommand{\t}{\mathbf{t}}
\newcommand{\te}{\mathbf{\tilde{t}}}

\newcommand{\FIDtx}{FID\textsubscript{ t \textrightarrow{} \~{y}}\xspace}
\newcommand{\FIDxt}{FID\textsubscript{ y \textrightarrow{} \~{t}}\xspace}

\newcommand{\SSIM}{SSIM}

\newcommand{\PG}{PatchGAN\xspace}
\newcommand{\IG}{ImageGAN\xspace}

\newcommand{\lsgan}{LSGAN \autocite{LSGAN}\xspace}
\newcommand{\hinge}{HINGE \autocite{lim2017geometric}\xspace}
\newcommand{\wgangp}{WGAN-GP \autocite{gulrajani2017improved}\xspace}

\newcommand{\Cut}{CUT \autocite{park2020cut}\xspace}
\newcommand{\CyG}{CycleGAN \autocite{CycleGAN2017}\xspace}
\newcommand{\TRB}{\emph{Turbo}\xspace}
\newcommand{\TRBU}{\emph{Turbo unpaired}\xspace}
\newcommand{\TRBP}{\emph{Turbo paired}\xspace}
\newcommand{\TRBDT}{\emph{Turbo digital twin}\xspace}
\newcommand{\UP}{UNET paired \autocite{Chaban2021wifs} \xspace}

\newcommand{\UN}{UNET\xspace}
\newcommand{\CNNG}{CNN-RESNET-CNN\xspace}

\newcommand{\Lpt}{\mathcal{L}_{\tilde{\mathbf{t}}}(\mathbf{t}, \tilde{\mathbf{t}})}
\newcommand{\Lpp}{\mathcal{L}_{\tilde{\mathbf{y}}}(\mathbf{y}, \tilde{\mathbf{y}})}

\newcommand{\Drp}{\mathcal{D}_{\mathbf{y} \hat{\mathbf{y}}}(\hat{\mathbf{y}})}
\newcommand{\Det}{\mathcal{D}_{\bf{t } \tilde{\mathbf{t}}}(\tilde{\mathbf{t}})}

\newcommand{\Drt}{\mathcal{D}_{\bf{t}\hat{\mathbf{t}}}(\hat{\mathbf{t}})}
\newcommand{\Dep}{\mathcal{D}_{\bf {y } \tilde{\mathbf{y}}}(\tilde{\mathbf{y}})}

\def\px_z{p({\bf x}|{\bf z})}

\def\pc_z{p({\bf c}|{\bf z})}

\def\IXZ{I({\bf X}; {\bf Z})}
\def\IXZ_phi{I_{\boldsymbol \phi}({\bf X}; {\bf Z})}

\def\pz_x{p({\bf z}|{\bf x})}

\def\pthx_z{p_{\boldsymbol \theta}({\bf x}|{\bf z})}

\def\px{\mathit{p}_{\mathrm{x}}(\bf{x})}

\def\pz{\mathit{p}_{\mathrm{z}}(\bf{z})}
\def\pc{\mathit{p}_{\mathrm{c}}(\bf{c})}

\def\qty{\mathit{q}_{\phi_{\mathrm t}}(\mathbf{t} | \mathbf{y})}

\def\x{\bf x}
\def\t{\bf t}
\def\y{\bf y}

\let\autocite\cite

\begin{document}
\bstctlcite{MyBSTcontrol} %
\renewcommand{\figureautorefname}{Fig.\negthinspace}

\title{Digital twins of physical printing-imaging channel \\
    \thanks{S. Voloshynovskiy is a corresponding author.}
    \thanks{This research was partially funded by the Swiss National Science Foundation SNF No. 200021\_182063.}
}

\author{\IEEEauthorblockN{Yury Belousov, Brian Pulfer, Roman Chaban, Joakim Tutt, Olga Taran, Taras Holotyak and Slava Voloshynovskiy}
	\IEEEauthorblockA{\textit{Department of Computer Science, University of Geneva, Switzerland} \\
		\{yury.belousov, brian.pulfer, roman.chaban, joakim.tutt, olga.taran, taras.holotyak, svolos\}@unige.ch}
}

\maketitle

\begin{abstract}
In this paper, we address the problem of modeling a printing-imaging channel built on a machine learning approach a.k.a. {\em digital twin} for anti-counterfeiting applications based on copy detection patterns (CDP). The digital twin is formulated on an information-theoretic framework called {\em Turbo} that uses variational approximations of mutual information developed for both encoder and decoder in a two-directional information passage. The proposed model generalizes several state-of-the-art architectures such as adversarial autoencoder (AAE) \cite{makhzani2015adversarial}, \CyG and adversarial latent space autoencoder (ALAE) \cite{pidhorskyi2020adversarial}. This model can be applied to any type of printing and imaging and it only requires training data consisting of digital templates or artworks that are sent to a printing device and data acquired by an imaging device. Moreover, these data can be paired, unpaired or hybrid paired-unpaired which makes the proposed architecture very flexible and scalable to many practical setups. We demonstrate the impact of various architectural factors, metrics and discriminators on the overall system performance in the task of generation/prediction of printed CDP from their digital counterparts and vice versa. We also compare the proposed system with several state-of-the-art methods used for image-to-image translation applications. The code and extended results of the simulation are publicly \href{https://gitlab.unige.ch/Yury.Belousov/digital-twin}
{available}\footnote{
\url{https://gitlab.unige.ch/sip-group/digital-twin} 
}.
\end{abstract}


\section{Introduction}

In recent years copy detection patterns (CDP) \cite{picard2004digital, voloshynovskiy2016physical} attracted a lot of attention as an anti-counterfeiting technology. At the same time, a lot of research was done to investigate the different factors impacting the authentication accuracy of CDP. However, the production of datasets of real CDP is a costly and timely process. It requires the printing and acquisition of original CDP and the production and acquisition of fakes, preferably on equipment close to the industrial one. The factors of cost, time and needed domain knowledge considerably constrain the study of the anti-counterfeiting aspects of CDP. 

The lack of accurate mathematical models of complex production and acquisition systems leads to a need to collect a huge amount of data for each particular case, reduces the system scalability to new products, production technologies and imaging devices, and makes the optimization process difficult, time-consuming and expensive. Moreover, the optimization of this system is complicated by a non-differentiable nature of existing models and their non-stochastic nature that does not reflect real practical situations.

The knowledge of the physical printing-imaging channel plays a very important role in anti-counterfeiting systems and is crucial for both the defender and the attacker. On the side of the defender, the knowledge of a model for this channel can 
(a) enable the overall optimisation of the whole authentication system by end-to-end training of encoders, decoders and decision modules, 
(b) simulate and predict the intra-class variabilities and 
(c) help generate synthetic samples of both originals and fakes that can be used to efficiently train decision module's classifiers.

The attacker can also benefit from such a model by (a) optimising the estimation of digital templates from the physical samples in the scope of copy attacks and (b) developing adversarial samples for the physical domain.

At the same time, the design of {\em digital twins} of printing-imaging channels is not a trivial task. To simplify it somehow, one can consider printing and imaging systems separately.

Besides some works \cite{nguyen2018probabilistic, villan2006multilevel} addressing the physics of specific production systems, there is no generalized theory on how to model even straightforward printing systems characterized by a high level of stochasticity and nonlinearity. The printing process model of each printing technology, such as off-set, digital off-set, inkjet or flexo, representing the most significant interest for practical applications, is very complex and domain-specific. Moreover, such a model should consider not only hardware but also software particularities of drivers that significantly impact the printed outcome. Altogether, it requires a lot of domain-specific know-how and makes the model development for each printing system very time-consuming. Furthermore, the validation of the model is also expensive and might require tuning many parameters.

\begin{figure*}[t!]
	\centering
	\includegraphics[width=0.8\linewidth]{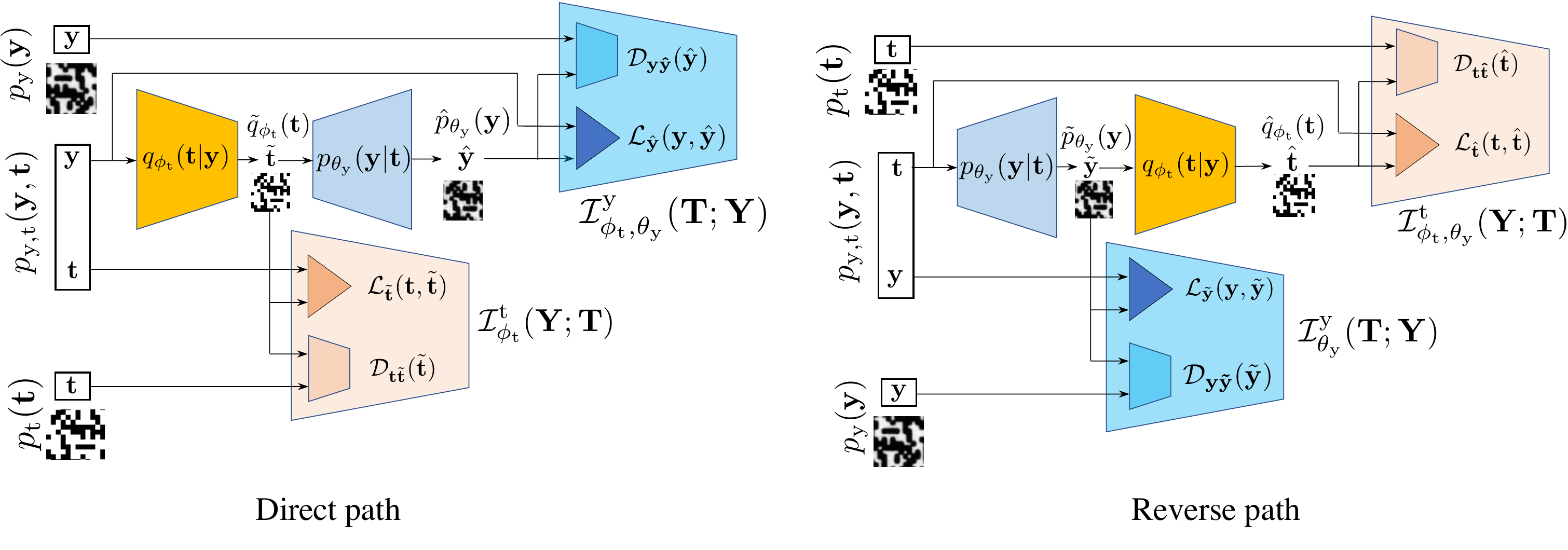}
	\caption{\TRBDT system: direct and reverse paths.} %
	\label{fig:basic_concep}
\end{figure*}

Not less important is the modelling of the acquisition/imaging process. Besides some remarkable exceptions  \cite{azzari2014gaussian, foi2008practical} that
present the models of noise in the CCD and CMOS imaging devices and practical methodologies of their validation, the simulation of the interaction between the incident light and reflecting object surface is not a trivial task \cite{ wong2015counterfeit}. The imaging device hardware components and drivers' settings such as type of sensor, resolution of sensors, optics, ISO, shutter time,  denoising, white color balancing, compression, etc., greatly impact the output image features. Finally, similarly to the models of production systems, there is no guarantee that the imaging model will be interpretable and differentiable and thus suitable for the envisioned ML tasks.

In this paper, we aim at addressing these challenges and shortcomings by following machine-learning framework and introducing a concept of {\em digital twins} of complex and unknown physical systems. More specifically, we  propose a \textit{digital twin} system that simulates the entire chain from the digital template $\bf t$ to the acquired image $\bf y$ that might represent both the original data $\bf x$ and fakes $\bf f$. The proposed system is based on an auto-encoder (AE) structure. To our best knowledge, there is no such framework for the addressed printing-imaging problem. 

The framework of \textit{digital twins} might be used as a simulator of complex physical printing-imaging systems for: 
\begin{itemize}
  \item creation of  differentiable models leading to the investigation of unexplored adversarial attacks in the physical world;
  \item generation of synthetic samples in order to train a supervised classifier for originals and fakes even when no fakes are known in advance by using synthetic samples as fakes;
  \item creation of augmentations for self-supervised learning (SSL) methods;
  \item investigation of variability in printing-imaging systems.
\end{itemize}

\noindent \textbf{Notations}  We use the following notations: $\t \in \{0, 1\}^{m \times m}$ denotes an original digital template; $\x \in [0, 1]^{m \times m}$ corresponds to an original printed code, while $\mathbf{f} \in [0, 1]^{m \times m}$ is used to denote a printed fake code; $\y \in [0, 1]^{m \times m}$ stands for a probe that might be either original or fake. We use $\mathbb{E}_{p(\x)}[.]$ to denote mathematical expectation with respect to a distribution $p(\x)$, $D_{\text{KL}}(.||.)$ denotes Kullback-Leibler divergence and $I(.;.)$ stands for mutual information. We assume that a pair of digital template $\t$ and CDP $\y$ are distributed as $(\y,  \t) \sim p_{\mathrm{y}, \mathrm{t}}(\y, \t)$.

\section{Proposed Turbo digital twin system}

The proposed \textit{digital twin} system is based on an auto-encoder structure and is represented by general stochastic encoder $\qty$ and decoder $p_{\theta_{\mathrm{y}}}(\mathbf{y}|\mathbf{t})$ that are deep networks parametrized by the parameters $\phi_{\mathrm{t}}$ and $\theta_{\mathrm{y}}$, respectively. The block diagram of \TRBDT system is shown in \autoref{fig:basic_concep}.

According to the proposed framework, given a pair of observable vectors $({\bf y}, {\bf t}) \sim p_{\mathrm{y}, \mathrm{t}}(\mathbf{y}, \mathbf{t})$, where ${\bf t}$ is a digital template and ${\bf y}$ is a printed code, i.e.\ either original ${\bf x}$ or fake ${\bf f}$, the system maximizes the mutual information between $\bf y$ and $\bf t$ for both encoder and decoder in direct and reverse paths.

We are considering a variational approximation for the {\em direct path} of the proposed system based on the maximization of two bounds on mutual information for the latent space and the reconstruction space:

\begin{equation}
    \begin{aligned}
	\mathcal{I}_{\phi_\mathrm{t} }^{\mathbf{t}}(\mathbf{Y} ; \mathbf{T})  
	& = \mathbb{E}_{p_{\mathrm{y}, \mathrm{t}}(\mathbf{y}, \mathbf{t})}\left[\log \frac{q_{\phi_\mathrm{t}}(\mathbf{t} | \mathbf{y})}{p_{\mathrm{t}}(\mathbf{t})} \frac{\tilde{q}_{\phi_\mathrm{t}}(\mathbf{t})}{\tilde{q}_{\phi_\mathrm{t}}(\mathbf{t})}\right] \\ 
	& \geq \underbrace{\mathbb{E}_{p_{\mathrm{y}}(\mathbf{y})} \mathbb{E}_{q_{\phi_\mathrm{t}}(\mathbf{t} | \mathbf{y})}\left[\log q_{\phi_\mathrm{t}}(\mathbf{t} | \mathbf{y})\right]}_{-\mathcal{L}_{\tilde{\mathbf{t}}}(\mathbf{t}, \tilde{\mathbf{t}})} \\
	& \;\;\; - \underbrace{D_{\mathrm{KL}}\left(p_{\mathrm{t}}(\mathbf{t}) \| \tilde{q}_{\phi_\mathrm{t}}(\mathbf{t})\right)}_{\mathcal{D}_{\mathbf{t} \tilde{\mathbf{t}}}(\tilde{\mathbf{t}})},
	\end{aligned}
	\label{eq:direct path1}
\end{equation}

\begin{equation}
    \begin{aligned}
	\mathcal{I}_{\phi_\mathrm{t}, \theta_\mathrm{y}
	}^{\mathbf{y}}(\mathbf{T} ; \mathbf{Y}) 
	& =  \mathbb{E}_{p_{\mathbf{y}, \mathbf{t}}(\mathbf{y}, \mathbf{t})}\left[\log \frac{p_{\theta_\mathrm{y}}(\mathbf{y} | \mathbf{t})}{p_{\mathbf{y}}(\mathbf{y})} \frac{\hat{p}_{\theta_\mathrm{y}}(\mathbf{y})}{\hat{p}_{\theta}(\mathbf{y})}\right] \\
	& \geq \underbrace{\mathbb{E}_{p_{\mathbf{y}}(\mathbf{y})} \mathbb{E}_{q_{\phi_\mathrm{t}
		}(\mathbf{t} | \mathbf{y})}\left[\log p_{\theta_\mathrm{y}}(\mathbf{y} | \mathbf{t})\right]}_{-\mathcal{L}_{\hat{\mathbf{y}}}(\mathbf{y}, \hat{\mathbf{y}})} \\
    & \;\;\; - \underbrace{D_{\mathrm{KL}}\left(p_{\mathbf{y}}(\mathbf{y}) \| \hat{\mathbf{t}}_{\theta}(\mathbf{y})\right)}_{\mathcal{D}_{\mathbf{y} \hat{\mathbf{y}}}(\hat{\mathbf{y}})}.
	\end{aligned}
	\label{eq:direct path2}
\end{equation}

Thus, the network is trained in such a way to maximise a weighted sum  of  (\ref{eq:direct path1}) and (\ref{eq:direct path2}) in order to find the best parameters $\phi_{\mathrm t}$ and $\theta_{\mathrm y}$ of the encoder and the decoder, respectively. This is achieved in the direct path by minimising the $\overline{\mathcal{L}}^{\text {Direct }}$ loss, representing the left network shown in \autoref{fig:basic_concep}:

\begin{equation}
    \begin{aligned} 
	\overline{\mathcal{L}}^{\text {Direct }}(\phi_{\mathrm t}, \theta_{\mathrm y}) 
	& =  \mathcal{L}_{\tilde{\mathbf{t}}}(\mathbf{t}, \tilde{\mathbf{t}})+\mathcal{D}_{\bf{t } \tilde{\mathbf{t}}}(\tilde{\mathbf{t}}) \\
	 &+\alpha \mathcal{L}_{\hat{\mathbf{y}}}(\mathbf{y}, \hat{\mathbf{y}})+\alpha \mathcal{D}_{\mathbf{y} \hat{\mathbf{y}}}(\hat{\mathbf{y}}),
	\end{aligned} 
	\label{eq:direct path3}
\end{equation}
where $\alpha$ is a parameter controlling the trade-off between the terms (\ref{eq:direct path1}) and (\ref{eq:direct path2}).

The variational approximation for the {\em reverse path} is:

\begin{equation}
    \begin{aligned} 
	\mathcal{I}_{\theta_{\mathrm{y}}}^{\mathbf{y}}(\mathbf{T} ; \mathbf{Y}) 
	& \geq \underbrace{\mathbb{E}_{p_{\mathrm t}(\mathbf{t})} \mathbb{E}_{p_{\theta_{\mathrm{y}}}(\mathbf{y} | \mathbf{t})}\left[\log p_{\theta_{\mathrm{y}}}(\mathbf{y} | \mathbf{t})\right]}_{-\mathcal{L}_{\tilde{\mathbf{y}}}(\mathbf{y}, \tilde{\mathbf{y}})} \\
	& \;\;\; - \underbrace{D_{\mathrm{KL}}\left(p_{\mathbf{y}}(\mathbf{y}) \| \tilde{p}_{\theta_{\mathrm{y}}}(\mathbf{y})\right)}_{\mathcal{D}_{\mathbf{y} \tilde{\mathbf{y}}}(\tilde{\mathbf{y}})},
	\end{aligned} 
	\label{eq:reverse path1}
\end{equation}

\begin{equation}
    \begin{aligned} 
	\mathcal{I}_{\phi_{\mathrm{t}}, \theta_{\mathrm{y}}}^{\mathbf{t}}(\mathbf{Y} ; \mathbf{T}) 
	& \geq \underbrace{\mathbb{E}_{p_{{\mathrm{t}}}(\mathbf{t})} \mathbb{E}_{p_{\theta_{\mathrm{y}}}(\mathbf{y} | \mathbf{t})}\left[\log q_{\phi_{\mathrm{t}}}(\mathbf{t} | \mathbf{y})\right]}_{-\mathcal{L}_{\hat{\mathbf{t}}}(\mathbf{t}, \hat{\mathbf{t}})} \\
	& \;\;\; -  \underbrace{D_{\mathrm{KL}}\left(p_{\mathrm{t}}(\mathbf{t}) \| \hat{q}_{\phi_{\mathrm{t}}}(\mathbf{t})\right)}_{\mathcal{D}_{\mathrm{t} \hat{\mathrm{t}}}(\hat{\mathbf{t}})}.
	\end{aligned} 
	\label{eq:reverse path2}
\end{equation}
The reverse path loss $\overline{\mathcal{L}}^{\text {Reverse }}$, weighted by $ \beta$, is represented by the right network shown in \autoref{fig:basic_concep}:

\begin{equation}
    \begin{aligned}
	\overline{\mathcal{L}}^{\text {Reverse }}(\phi_{\mathrm t}, \theta_{\mathrm y}) 
	& =\mathcal{L}_{\tilde{\mathbf{y}}}(\mathbf{y}, \tilde{\mathbf{y}})+\mathcal{D}_{\bf {y } \tilde{\mathbf{y}}}(\tilde{\mathbf{y}}) \\
    &+\beta \mathcal{L}_{\hat{\mathbf{t}}}(\mathbf{t}, \hat{\mathbf{t}})+\beta \mathcal{D}_{\bf{t}\hat{\mathbf{t}}}(\hat{\mathbf{t}}).
	\end{aligned} 
	\label{eq:reverse path3}
\end{equation}

\section{Architectural details}
The \TRB system is flexible and allows different configurations. It can be used for paired data when all losses are preserved and we possess pairs of digital template $\bf t$ and CDP $\bf y$. In contrast, if such pairs are not available at the training that corresponds to the unpaired setup, the terms $\Lpt$ and $\Lpp$ disappear and one gets a \TRBU setup.

In addition, many existing models can be expressed as part of the \TRB framework. For example, the \CyG model can be obtained by removing the discriminators on reconstruction $\Drt$ and $\Drp$ from \TRBU. The pix2pix model \autocite{pix2pix2017} is also a part of the complete \TRB framework with the removed cycle losses while keeping  $\Lpt, \Det$ or $\Lpp, \Dep$ depending on the direction of training. The adversarial autoencoder (AAE)  
\cite{makhzani2015adversarial} corresponds to the direct path with the adversarial and reconstruction losses.
The CUT \autocite{park2020cut} and ALAE \cite{pidhorskyi2020adversarial} models can also be expressed through the Turbo framework. 

\subsection{Structure of encoder and decoder}
The proposed approach does not impose any restrictions on the encoder and decoder architecture, which allows a wide variety of options. In our work, we have considered several most widely used architectures for the encoders and decoders, namely:
\begin{itemize}
    \item \CNNG adapted from \CyG and StarGAN \autocite{choi2018stargan} models, consisting of two convolutional layers for downsampling, nine residual blocks \autocite{he2016deep}, and two transposed convolutional layers for upsampling.
    \item \UN \autocite{unet} with skip-connections layers.
\end{itemize}
In both cases, instance normalization \autocite{ulyanov2016instance} was used to stabilize training together with Adam optimizer \autocite{Adam}.

\subsection{Adversarial loss and structure of discriminator}
Selection of the adversarial loss, which implements $D_{\text{KL}}(.||.)$ terms, for the considered models could be crucial for the success of the training \autocite{lucic2018gans}. In our work, we examine three of the most popular losses: \lsgan, \hinge and WGAN \autocite{arjovsky2017wasserstein} with gradient penalty \autocite{gulrajani2017improved}.

We started with the standard PatchGAN \autocite{pix2pix2017} discriminator. However, we quickly discovered that in combination with a WGAN-GP loss, the results were extremely bad. We believe this follows from the fact that PatchGAN generates overlapping patches, which interfere 
when calculating the earth's moving distance. Therefore, we added another discriminator ``ImageGAN'', based on residual networks \autocite{he2016deep}, for the comparison, which takes the whole picture as the input and produces a single scalar output. 

\section{Training details}
We used PyTorch for all experiments.
One training cycle per model varies from one to four days using four RTX 2080 Ti or a single A100 80 GB card depending on the configuration. 

\subsection{Dataset}
For the empirical evaluation of the proposed \TRB framework, we use the Indigo 1x1 base dataset \autocite{Chaban2021wifs} that consists of CDP with $1 \times 1$ pixel symbol size. This dataset contains 720 samples that we divide at 80\% and 20\% for the training and test sets, respectively. For the sake of experimental purity, the same non-intersecting split is used in all trials. To speed up the study, each original image of size $684 \times 684$ pixels is divided into four non-overlapping crops of size $256 \times 256$ each. Due to the paper length limit, all of the results below are obtained for the HP Indigo 7600 printer (HPI 76), but we do not observe significant differences when codes printed on another printer are used as input.

\subsection{Setups under consideration}
To our best knowledge, all previous works in CDP field use only paired data for the estimation. However, we believe that this condition might not always hold. One of the key novelties of our work is that we consider the case where an attacker has an unpaired dataset, where there is no exact match between the digital template and the respective printed code, and all data are represented as an unordered set.

However, the flexibility of the \TRB framework allows the use of paired losses $\Lpt$ and $\Lpp$ if paired data is available. It is also possible to train only one path estimation for example from the template to the printed code or vice versa.

\subsection{Stability of training} \label{section:hacks}
Adversarial training with discriminators is known to be quite unstable due to the mode collapse and vanishing gradients. Therefore, the following refinements were investigated to improve the quality of results:
\begin{itemize}
    \item Balancing discriminator and generator iterations via the number of discriminator iterations per generator iteration $n_{D}$ \autocite{gulrajani2017improved}.
    \item However, selecting the appropriate number of iterations is not an obvious task. Therefore, in the case of constraints on the possible values of loss function, i.e., in the case of LSGAN --- values are non-negative, instead of one parameter, a principled approach is preferable, where the discriminator is updated if its loss is greater than $D_{\text{threshold}}$ (discriminator poorly separates the generated samples) or the generator's loss is less than $G_{\text{threshold}}$ (the generated samples easily fools the discriminator) \autocite{gan_training}.
    \item Updating the discriminator using the history of generated images, rather than only those generated at the last iteration
    \autocite{image_pool_shrivastava2017learning}.
    \item Flipping labels from time to time when training the discriminator with probability $p_{\text{flip}}$ \autocite{gan_training} and adding some artificial noise to the discriminator's inputs \autocite{noise_sonderby2016amortised} with probability $p_{\text{noise}}$ and weight $w_{\text{noise}}$. We have experimented with ways of combining these two heuristics and noticed that together they give better results compared to using only one or none of them.
\end{itemize}

\section{Computer simulation}
The reported results are obtained without any post-processing and represent a direct output of deep networks. Additional post-processing 
might increase the accuracy of digital template estimation and generation. However, to preserve the scalability to any artwork and fair comparison, we report all results without any refinements.
The UNET paired model from \autocite{Chaban2021wifs} is used as a baseline.

\subsection{Metrics}
The following metrics were used to evaluate the quality of the predictions:
\begin{itemize}
    \item Hamming distance $d_H(\t, \operatorname{binary}(\te))$, where $\operatorname{binary}(.)$ is a binarization function. %
    \item Mean square error (MSE) distance $d_2(\y, \ye)$.
    \item Structural similarity index (\SSIM) $d_{\SSIM}(\y, \ye)$ introduced in \autocite{ssim_wang2004image} to address an issue that the mean squared error is not highly indicative of perceived similarity of images.
    \item Fréchet Inception Distance (FID): \FIDtx  and \FIDxt  proposed  in \autocite{NIPS2017_FID}. Instead of a simple pixel-by-pixel comparison of images, $\operatorname{FID}$ estimates the mean and standard deviation of one of the deep layers in the pretrained convolutional neural network. 
    We suppose that the usage of deep network statistics can be helpful not only as a measure of human perception of image similarity but also to assess the difficulty of distinguishing the generated images from the real ones since the network activations are similar at a metric close to zero.
\end{itemize}

\subsection{Evaluation}

First of all, we investigated the impact of the encoder-decoder architecture. 
The obtained results are shown in \autoref{table:turbo-backbone}. In all scenarios, the configuration with \CNNG performs better than with \UN. However, in the case of paired data, the difference is less significant. 
The \TRB paired also outperforms CycleGAN
with respect to most metrics and is also less sensitive to the choice of architecture.

{
\renewcommand{\arraystretch}{1.5}
\sisetup{detect-mode,round-mode=places,detect-weight=true, round-precision=3}

\begin{table*}
	\centering
	\caption{TURBO performance with regard to the encoder-decoder backbone}
    \label{table:turbo-backbone}
	\begin{tabular}{ccSSSSSS}
		\toprule
		\textbf{Model} & \textbf{backbone} & \textbf{\FIDxt} & \textbf{Hamming distance} & \textbf{\FIDtx} & \textbf{MSE} & \textbf{\SSIM} \\
		\midrule

        \multirow{2}{*}{\CyG}  & \UN           & 9.5992 & 0.1946 & 12.3991 & 0.0644 & 0.6753 \\
                       & \CNNG & 3.8653 & 0.1549 & \B 4.4507  & 0.0490 & 0.7315 \\ \midrule
\multirow{2}{*}{\TRBP} & \UN           & 4.272 & 0.1072   & 8.9417  & 0.0428 & 0.7654 \\
                       & \CNNG & \B 3.164 & \B 0.0855 & 6.6049  & \B 0.0400 & \B 0.7787 \\

		\bottomrule
	\end{tabular}
\end{table*}

}

\autoref{table:turbo-combined} illustrates the impact of adversarial loss and discriminator type depending on the chosen \TRB configuration.  It should be noted that the configuration with \wgangp  does not converge when used together with \PG, but shows one of the best results with \IG.
{
\renewcommand{\arraystretch}{1.5}
\sisetup{detect-mode,round-mode=places,detect-weight=true, round-precision=3}

\begin{table*}
	\centering
	\caption{TURBO performance with regard to the GAN loss and discriminator type}
    \label{table:turbo-combined}
	\begin{tabular}{cccSSSSSS}
		\toprule
		\textbf{Model}                   & \textbf{GAN Loss}           & \textbf{Discriminator type} & \textbf{heuristics}\textsubscript{\ref{section:hacks}} & \textbf{\FIDxt}  & \textbf{Hamming distance} & \textbf{\FIDtx}  & \textbf{MSE}     & \textbf{\SSIM} \\ \midrule

        \multirow{12}{*}{\CyG}  & \multirow{4}{*}{\lsgan}  & \multirow{2}{*}{\PG} & \xmark & 66.9678  & 0.2135 & 14.6107 & 0.0611 & 0.6885 \\
                        &                          &                      & \cmark & 46.2153  & 0.2053 & 9.8276  & 0.0638 & 0.6834 \\
                        &                          & \multirow{2}{*}{\IG} & \xmark & 48.8007  & 0.1983 & 12.4811 & 0.0654 & 0.6681 \\
                        &                          &                      & \cmark & 65.4581  & 0.2031 & 5.7128  & 0.0616 & 0.6944 \\ \cmidrule{2-9}
                        & \multirow{4}{*}{\hinge}  & \multirow{2}{*}{\PG} & \xmark & 4.0735   & 0.1944 & \B 4.4507  & 0.0654 & 0.6728 \\
                        &                          &                      & \cmark & 4.7414   & 0.1843 & 9.3397  & 0.0616 & 0.6873 \\
                        &                          & \multirow{2}{*}{\IG} & \xmark & 124.9961 & 0.2432 & 20.2538 & 0.0705 & 0.6504 \\
                        &                          &                      & \cmark & 28.2055  & 0.1923 & 5.9783  & 0.0659 & 0.6737 \\ \cmidrule{2-9}
                        & \multirow{4}{*}{\wgangp} & \multirow{2}{*}{\PG} & \xmark & 236.8354 & 0.2254 & 68.5791 & 0.0785 & 0.6279 \\
                        &                          &                      & \cmark & 248.5826 & 0.2310 & 80.3620 & 0.0776 & 0.6278 \\
                        &                          & \multirow{2}{*}{\IG} & \xmark & 3.8653   & 0.1549 & 8.3979  & 0.0574 & 0.7136 \\
                        &                          &                      & \cmark & 4.1301   & 0.1628 & 17.4147 & 0.0490 & 0.7315 \\ \midrule
\multirow{10}{*}{\TRBU} & \multirow{4}{*}{\lsgan}  & \multirow{2}{*}{\PG} & \xmark & 33.3027  & 0.2006 & 12.2998 & 0.0630 & 0.6737 \\
                        &                          &                      & \cmark & 51.8154  & 0.2084 & 14.3994 & 0.0615 & 0.6809 \\
                        &                          & \multirow{2}{*}{\IG} & \xmark & 136.3163 & 0.2105 & 20.8801 & 0.0698 & 0.6674 \\
                        &                          &                      & \cmark & 26.1511  & 0.1951 & 12.0478 & 0.0646 & 0.679  \\ \cmidrule{2-9}
                        & \multirow{4}{*}{\hinge}  & \multirow{2}{*}{\PG} & \xmark & 4.4791   & 0.1947 & 12.6512 & 0.0634 & 0.6735 \\
                        &                          &                      & \cmark & 3.5713   & 0.1864 & 16.1211 & 0.0644 & 0.6781 \\
                        &                          & \multirow{2}{*}{\IG} & \xmark & 322.3736 & 0.4324 & 54.0779 & 0.1889 & 0.3    \\
                        &                          &                      & \cmark & 25.7168  & 0.1771 & 16.8364 & 0.0589 & 0.6828 \\ \cmidrule{2-9}
                        & \multirow{2}{*}{\wgangp} & \multirow{2}{*}{\IG} & \xmark & 3.601    & 0.155  & 5.9117  & 0.0646 & 0.6783 \\
                        &                          &                      & \cmark & 4.3332   & 0.1672 & 15.031  & 0.0636 & 0.6849 \\ \midrule
\multirow{2}{*}{\TRBP} & \multirow{2}{*}{\wgangp} & \multirow{2}{*}{\IG} & \xmark & \B 3.164 & \B 0.0855 & 6.6049 & 0.0430 & 0.772 \\
                       &                          &                      & \cmark & 4.3120 & 0.0917 & 9.7115 & \B 0.0400 & \B 0.7787 \\
        
		\bottomrule
	\end{tabular}
\end{table*}

}

The best results among all investigated configurations are summarized in \autoref{table:final}. It is obvious that the models without pairwise information perform worse. The \TRB configurations outperform also contrastive system based on the CUT model, and the \TRBP outperforms the baseline in almost all metrics.
{
\renewcommand{\arraystretch}{1.5}
\sisetup{detect-mode,round-mode=places,detect-weight=true, round-precision=4}

\begin{table*}
	\centering
	\caption{performance with regard to the model}
    \label{table:final}
	\begin{tabular}{cSSSSSS}
		\toprule
		\textbf{Model} & \textbf{\FIDxt} & \textbf{Hamming distance} & \textbf{\FIDtx} & \textbf{MSE} & \textbf{\SSIM} \\
		\midrule

\Cut   & 3.8644  & 0.1990 & 5.2941  & 0.0610 & 0.6979 \\
\CyG                          & 3.8653  & 0.1549 & \B 4.4507  & 0.0490 & 0.7315 \\
\TRBU                         & 3.5713  & 0.1550 & 5.9117  & 0.0589 & 0.6849 \\
\TRBP                         & \B 3.164  & \B 0.0855 & 6.6049  & 0.0400 & \B 0.7787 \\
\UP                           & 6.2113  & 0.1002 & 28.1099 & \B 0.0363 & 0.7775 \\

		\bottomrule
	\end{tabular}
\end{table*}

} 

{
\renewcommand{\arraystretch}{1.4}

\newcommand{\sample}{000651_2.png}

\begin{table*}[t!]
	\centering
	\caption{Examples of synthetic samples generated by several studied systems: \\ the top row shows the estimated digital templates $\te$,
 the bottom row visualises the printing estimations $\ye$ \\
 from their real counterparts in the first column}
	\label{table:samples}
	\begin{tabular} {
			 c
			 c
			 c
			 c
			 c
			 c}
		\toprule

		\textbf{original}                                                                                      & \textbf{\Cut}                                                                                      & \textbf{\CyG}                                                                                      & \textbf{\TRBU}                                                                                      & \textbf{\TRBP}                                                                                      & \textbf{\UP}                                                                                      \\ \midrule
        
    \includegraphics[width=0.14\textwidth, keepaspectratio]{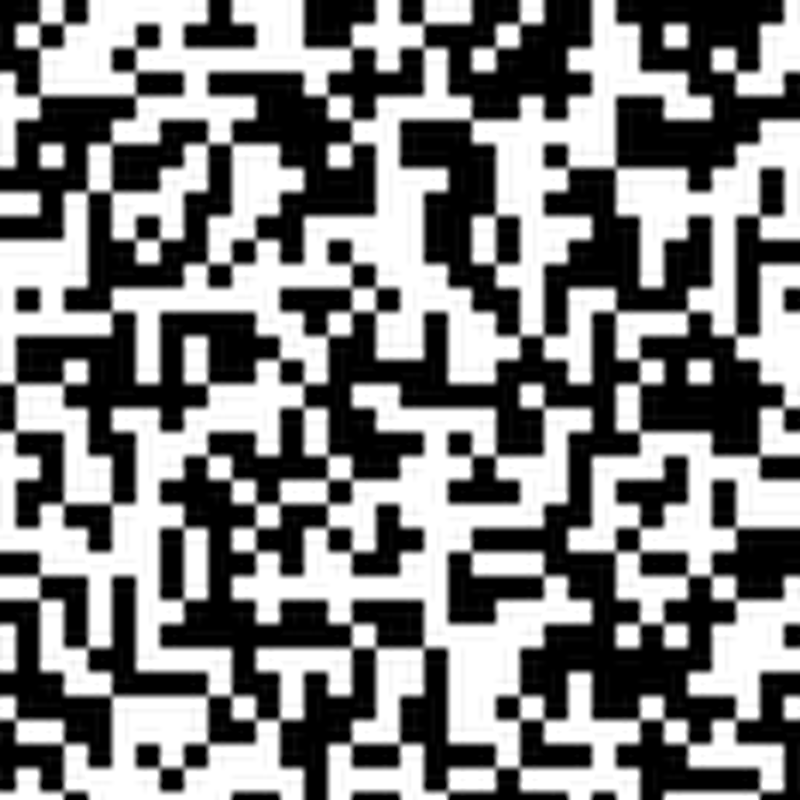} & \includegraphics[width=0.14\textwidth, keepaspectratio]{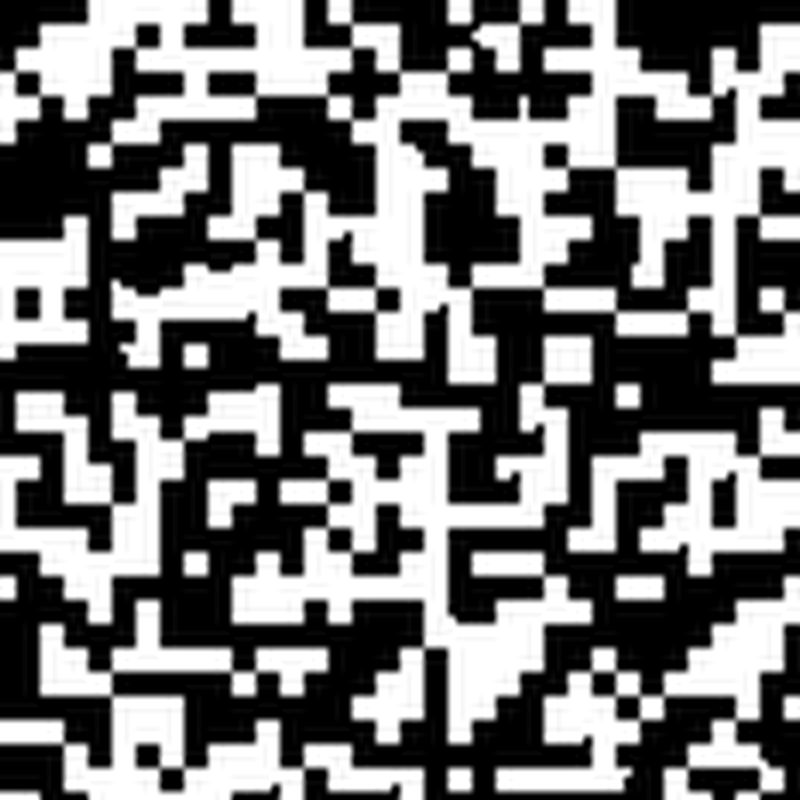} & \includegraphics[width=0.14\textwidth, keepaspectratio]{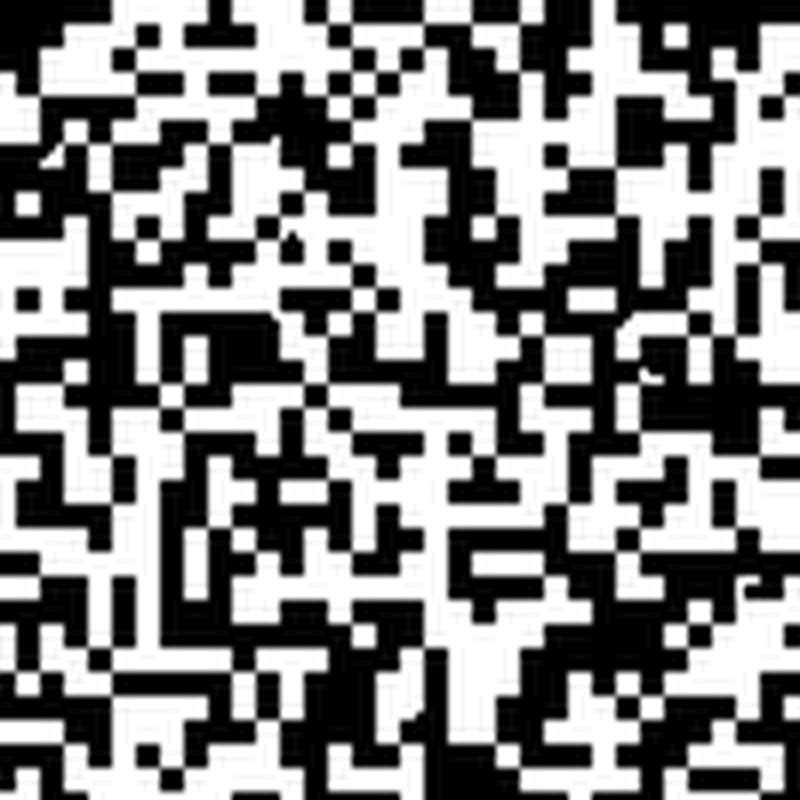} & \includegraphics[width=0.14\textwidth, keepaspectratio]{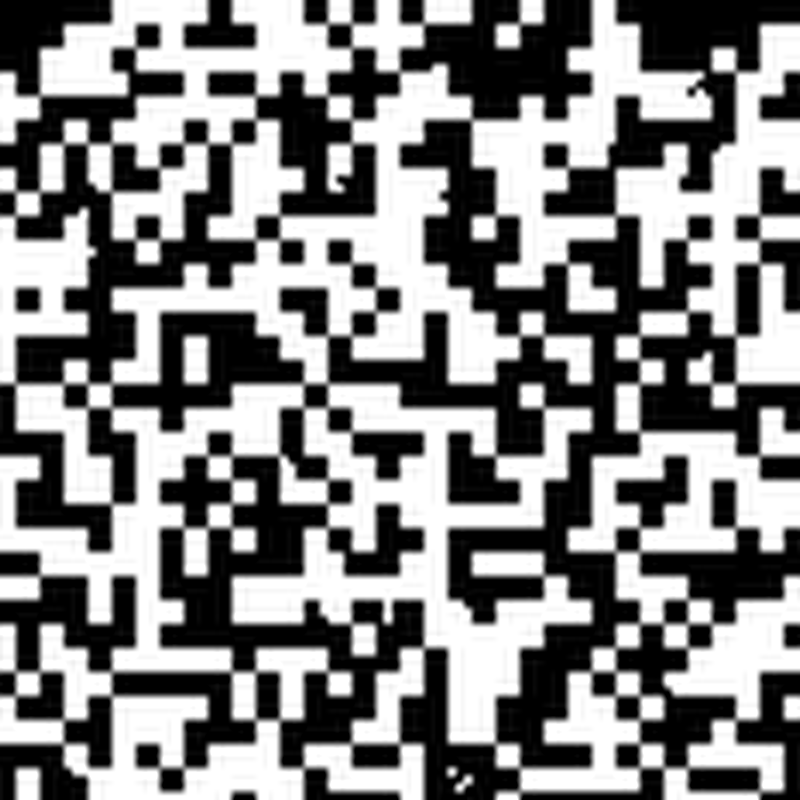} & \includegraphics[width=0.14\textwidth, keepaspectratio]{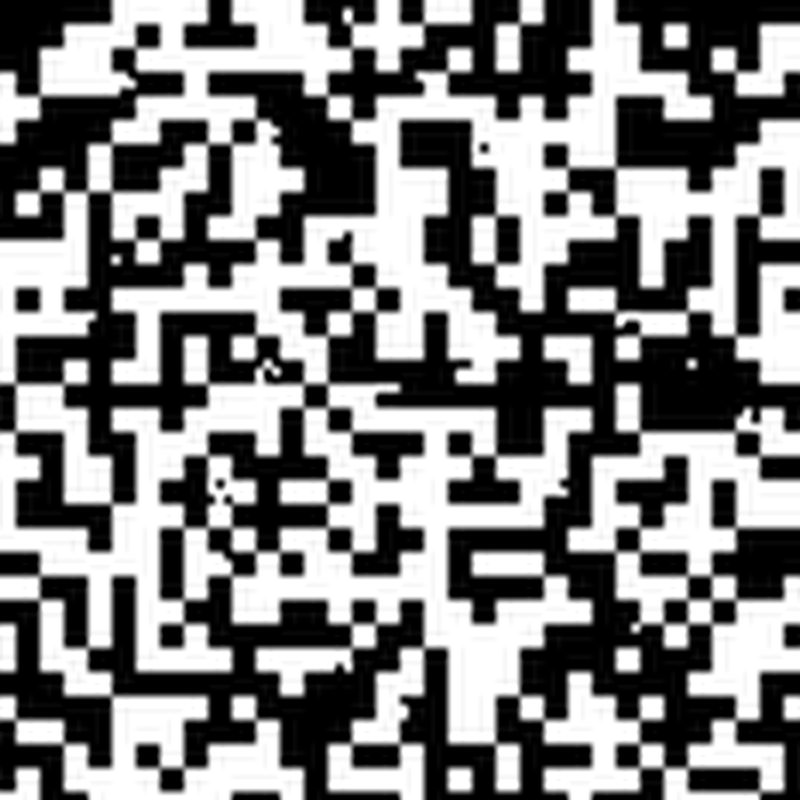} & \includegraphics[width=0.14\textwidth, keepaspectratio]{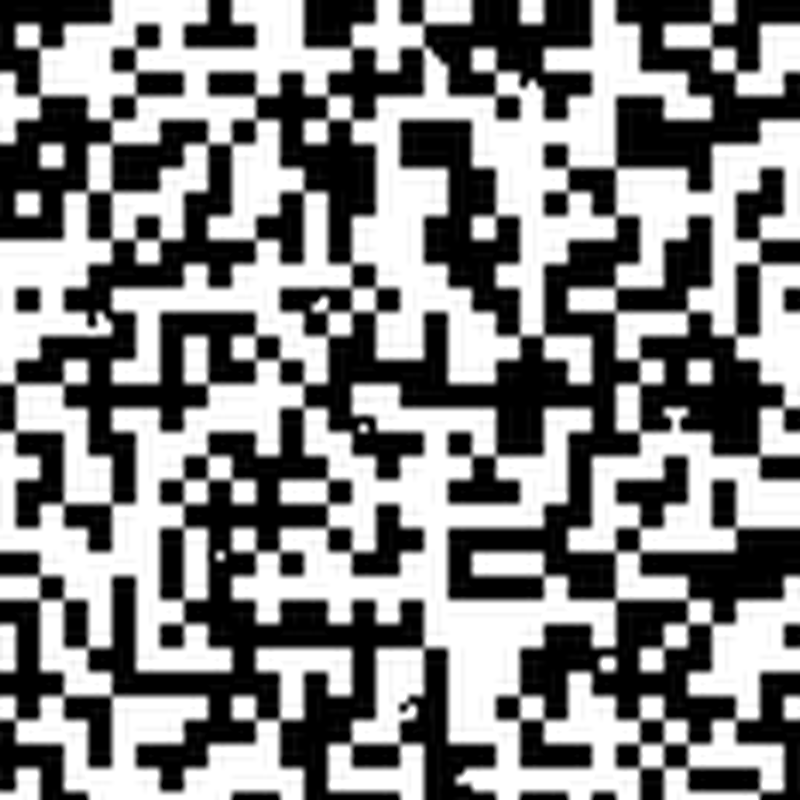} \\ 
    \midrule
    \includegraphics[width=0.14\textwidth, keepaspectratio]{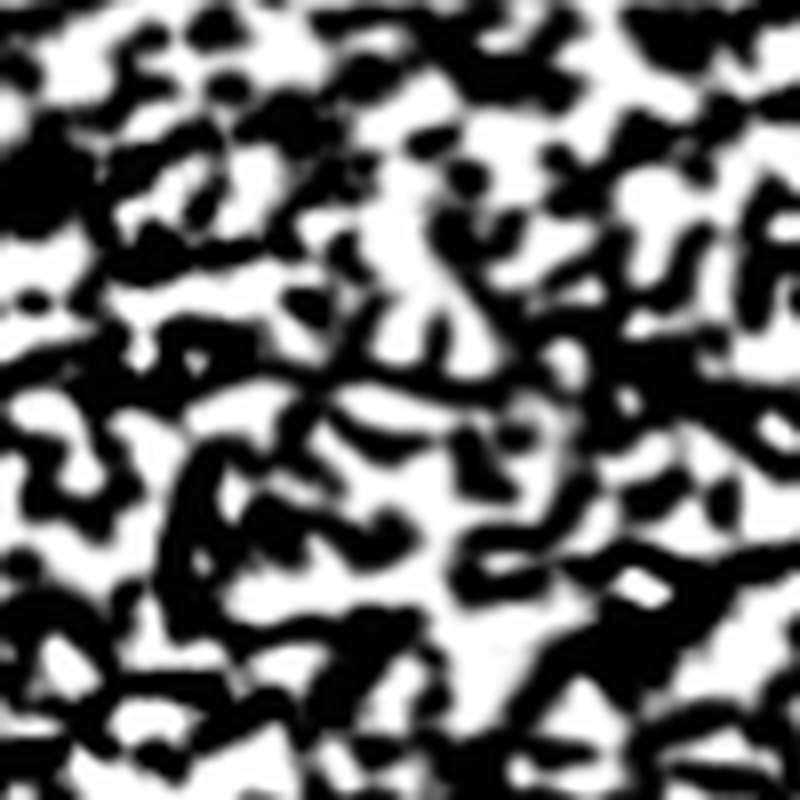} & \includegraphics[width=0.14\textwidth, keepaspectratio]{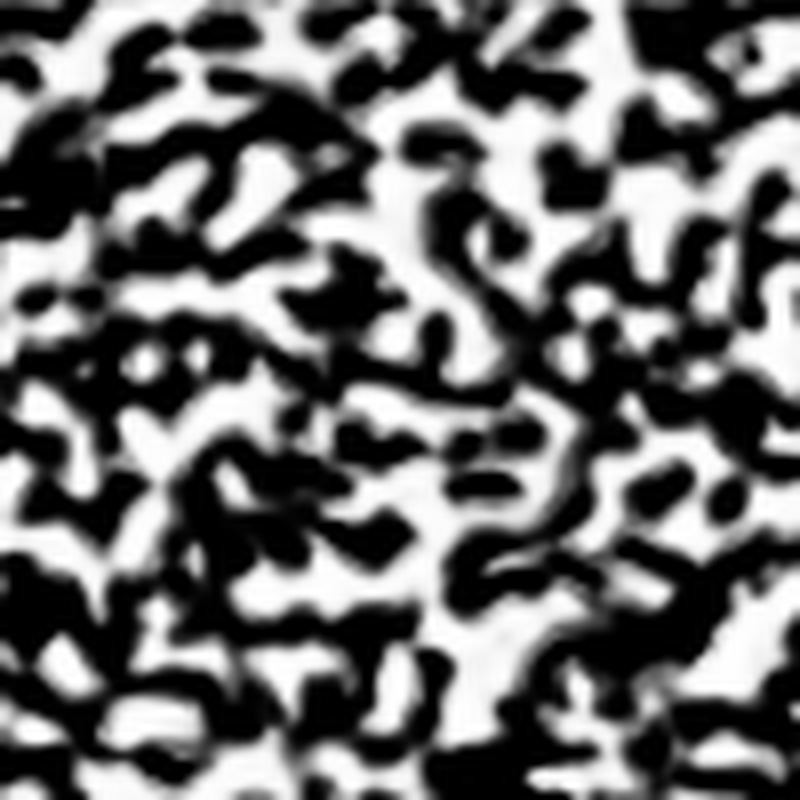} & \includegraphics[width=0.14\textwidth, keepaspectratio]{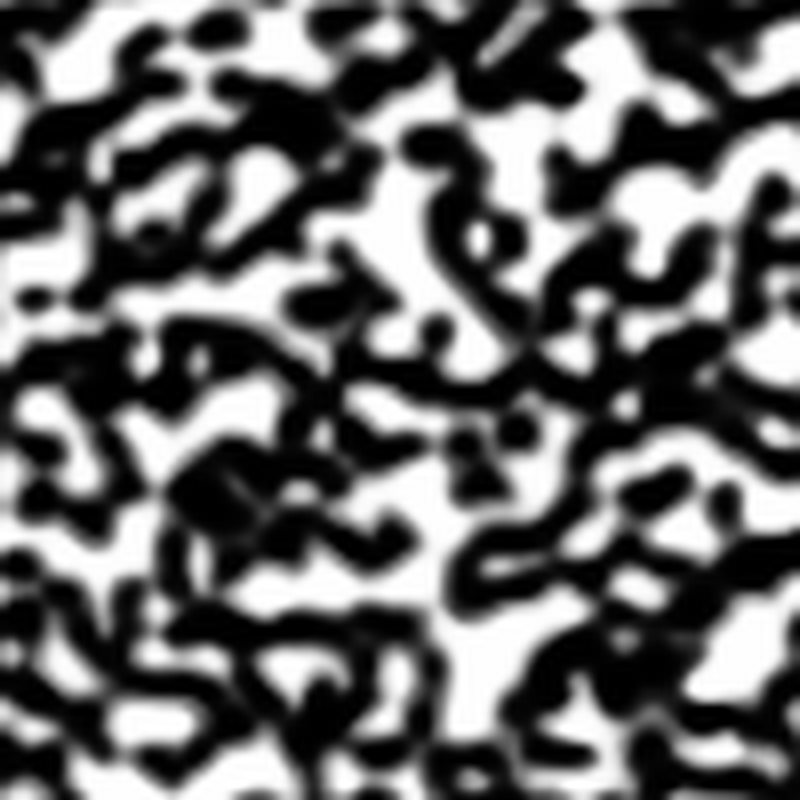} & \includegraphics[width=0.14\textwidth, keepaspectratio]{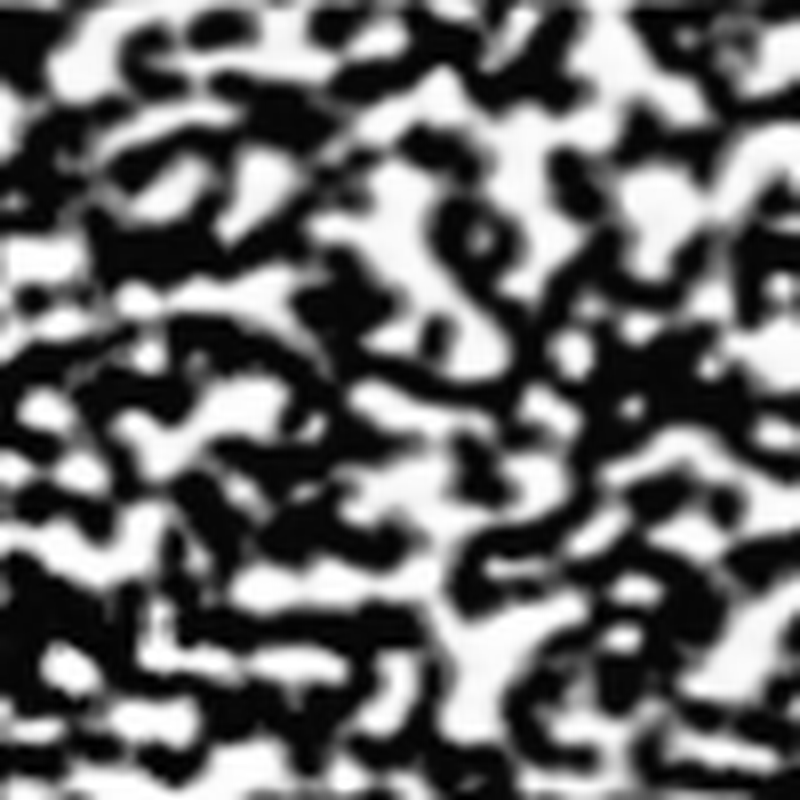} & \includegraphics[width=0.14\textwidth, keepaspectratio]{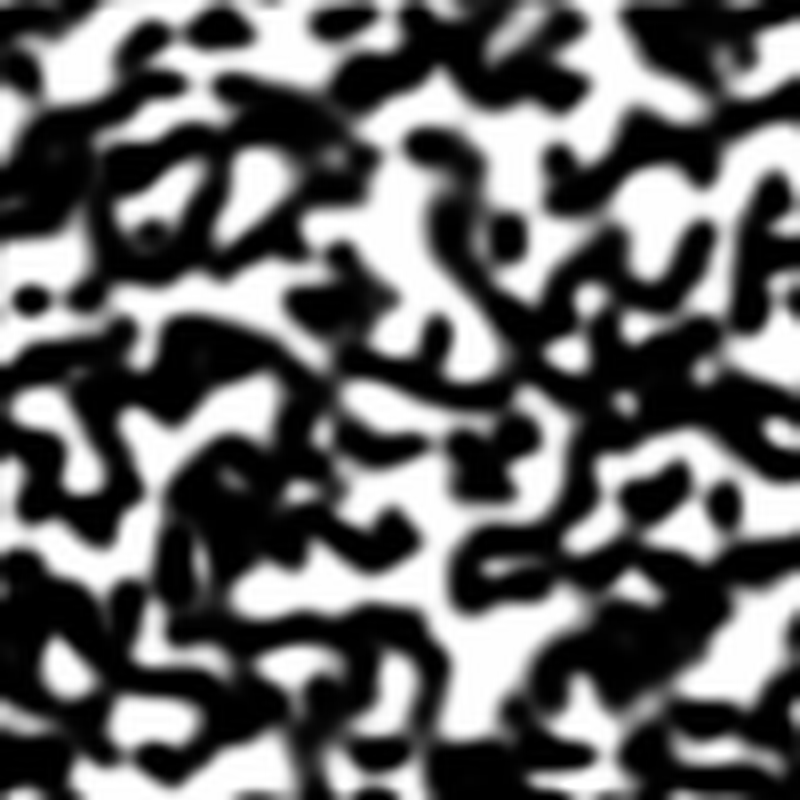} & \includegraphics[width=0.14\textwidth, keepaspectratio]{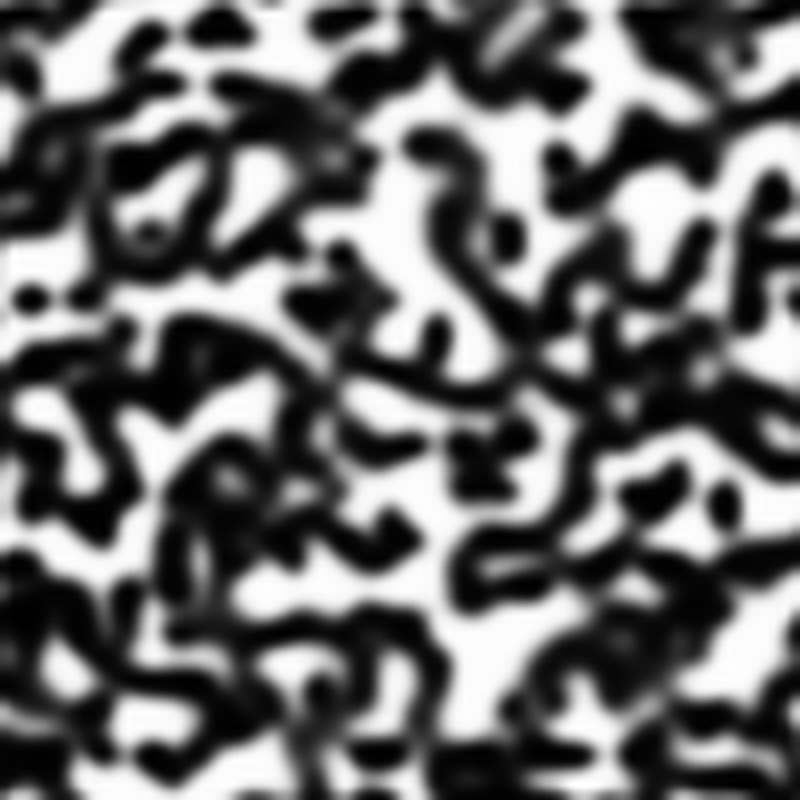} \\ 
    \bottomrule
    
	\end{tabular}
\end{table*}
}

\ifbool{umap_to_appendix}{}{
\begin{figure}[t!]
	\includegraphics[width=1.\linewidth]{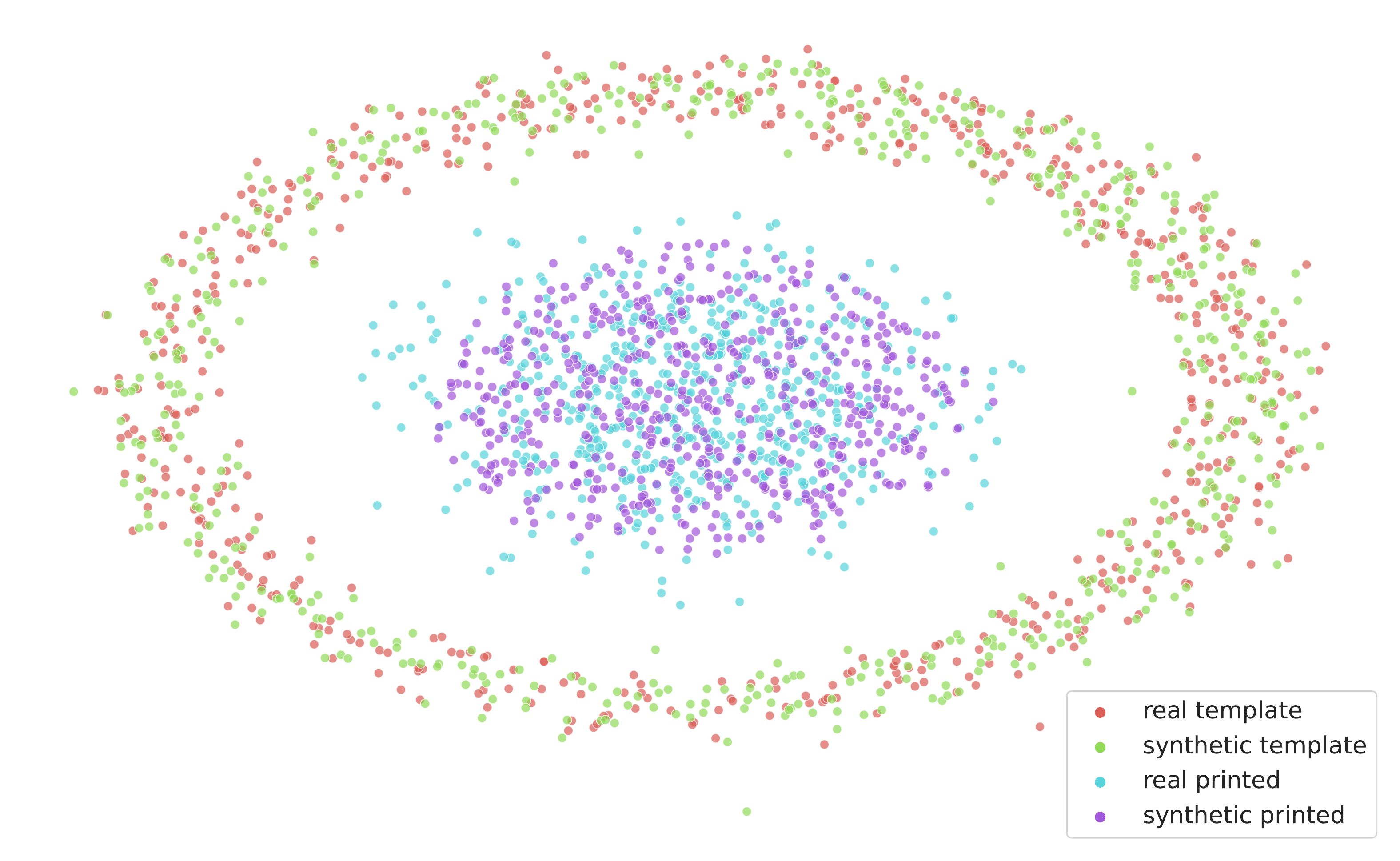}
	\caption{\label{fig:umap-vis}
 Umap \autocite{umap-soft} visualisation for artificially generated templates and printed jointly with digital templates and physically printed codes.}
\end{figure}

\subsection{Visualization}
To further assess the quality of the proposed models, an UMAP \autocite{umap-soft} 
visualisation was performed. The projection of artificially generated templates and printed codes from the test part jointly with corresponding original digital templates and physically printed codes is shown in \autoref{fig:umap-vis}. As expected, matching samples are close to each other, and there are different clusters for printed and template codes. 

}

\subsection{Visualization of synthetic samples}
To illustrate the quality of the synthetic samples produced by various systems studied in this paper, we pick up a random sample and show both synthetic digital templates estimated from physical CDP and vice versa in \autoref{table:samples}. Models that use paired examples show better generation performance, but models trained entirely in unpaired mode also perform decently. Visually, the synthetic samples look almost indistinguishable from their real counterparts.

\section{Conclusions}
In this paper, we present the \TRBDT framework for the simulation of the physical printing-imaging channel. We believe that such a differential model allows to consider the adversarial fakes for the physical world and also opens new perspectives for the optimization of authentication systems.

For future work, we will consider the usage of the generated examples to build a classifier based on the augmented synthetic samples of both original CDP and fakes. Additionally, issues of stochasticity and usage in hybrid settings, where only part of the data is paired, remain open for future research.

\bibliographystyle{IEEEtran}
\bibliography{bibliography}

\ifbool{umap_to_appendix}{
\clearpage \newpage
\appendix
\begin{figure}[t!]
	\includegraphics[width=1.\linewidth]{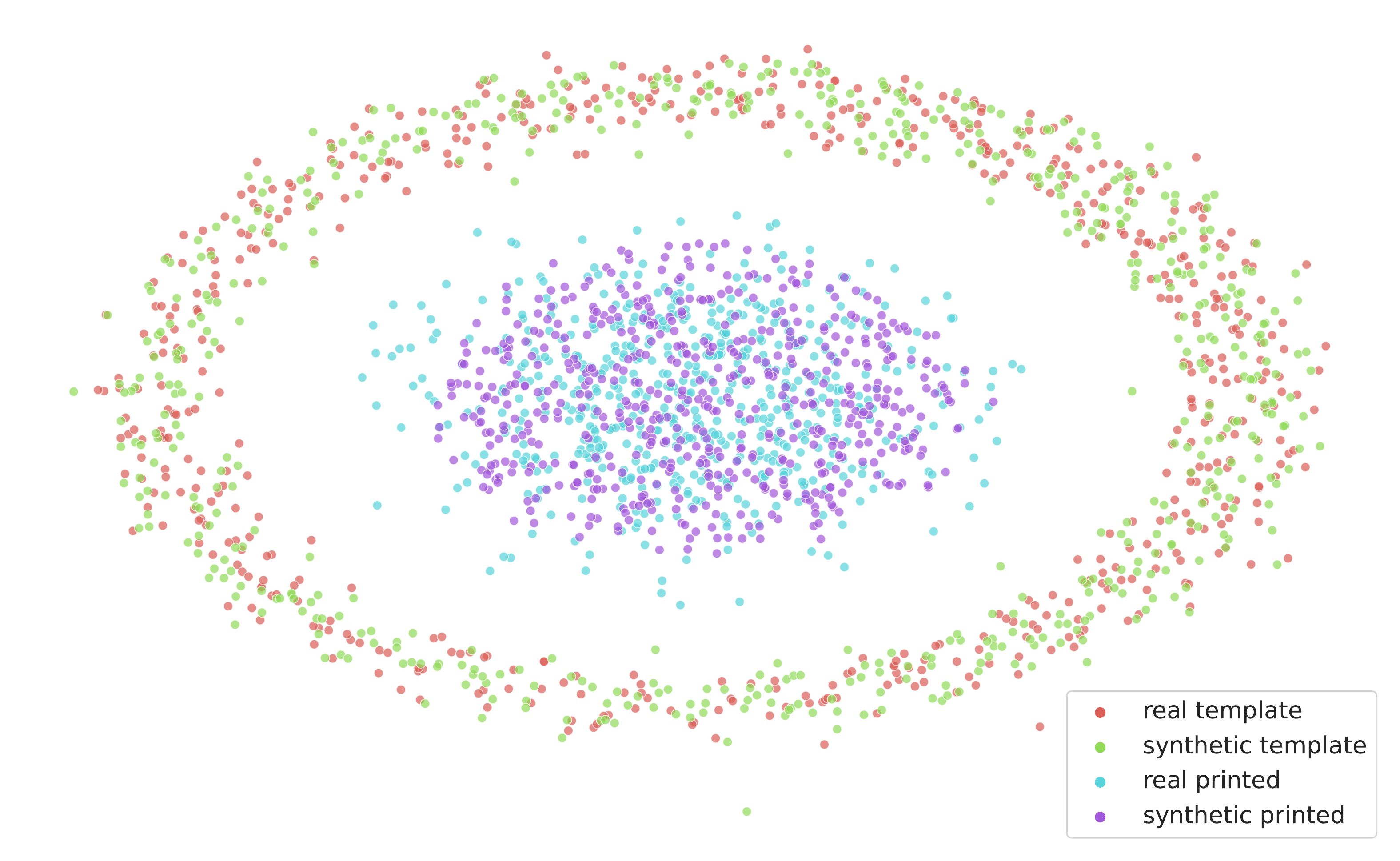}
	\caption{\label{fig:umap-vis}
 Umap \autocite{umap-soft} visualisation for artificially generated templates and printed jointly with digital templates and physically printed codes.}
\end{figure}

\subsection{Visualization}
To further assess the quality of the proposed models, an UMAP \autocite{umap-soft} 
visualisation was performed. The projection of artificially generated templates and printed codes from the test part jointly with corresponding original digital templates and physically printed codes is shown in \autoref{fig:umap-vis}. As expected, matching samples are close to each other, and there are different clusters for printed and template codes. 
}
{}

\end{document}